\documentclass[letterpaper, 10 pt, conference]{ieeeconf}

\IEEEoverridecommandlockouts                              

\usepackage{booktabs}
\usepackage{times}
\usepackage{epsfig}
\usepackage{graphicx}
\usepackage{amsmath}
\usepackage{amssymb}
\usepackage{svg}
\usepackage{multirow}
\graphicspath{{figure/}}
\usepackage[font=small,skip=0pt]{caption}
\usepackage{algorithm}
\usepackage{algpseudocode}
\usepackage{amsmath}
\usepackage{url,hyperref,xcolor}

\title{\LARGE \bf
ULSD: Unified Line Segment Detection across Pinhole, Fisheye, and Spherical Cameras
}

\author{Hao Li$^{*1}$, Huai Yu$^{*1,2}$, Wen Yang$^{1}$, Lei Yu$^{1}$ and Sebastian Scherer$^{2}$
\thanks{*Equal contribution.}
\thanks{$^{1}$Hao Li, Huai Yu, Wen Yang and Lei Yu are with the Electronic Information School, Wuhan University,  Wuhan 430072, China {\tt\small \{lihao2015, yuhuai, yangwen, ly.wd\}@whu.edu.cn}}%
\thanks{$^{2}$Sebastian Scherer is with the Robotics Institute, Carnegie Mellon University,
	Pittsburgh, PA 15213, USA
	{\tt\small basti@andrew.cmu.edu}}%
}

\begin{document}

\maketitle
\thispagestyle{empty}
\pagestyle{empty}

\begin{abstract}
	Line segment detection is essential for high-level tasks in computer vision and robotics. Currently, most state-of-the-art (SOTA) methods are dedicated to detecting straight line segments in undistorted pinhole images, thus distortions on fisheye or spherical images may largely degenerate their performance. Targeting at the unified line segment detection (ULSD) for both distorted and undistorted images, we propose to represent line segments with the Bezier curve model. Then the line segment detection is tackled by the Bezier curve regression with an end-to-end network, which is model-free and without any undistortion preprocessing. Experimental results on the pinhole, fisheye, and spherical image datasets validate the superiority of the proposed ULSD to the SOTA methods both in accuracy and efficiency (40.6fps for pinhole images). The source code is available at \href{https://github.com/lh9171338/Unified-Line-Segment-Detection}{https://github.com/lh9171338/Unified-Line-Segment-Detection}.  
	
\end{abstract}

\section{Introduction} \label{sec1}
Line segment detection is one of the most fundamental problems in computer vision and robotics, which can facilitate many high-level vision tasks such as image matching \cite{ImageMatching}, camera calibration \cite{FisheyeRectification3,FisheyeRectification4},  structure from motion \cite{SfM1,SfM2}, and visual SLAM \cite{StructVIO,vSLAM1,vSLAM2, yu2020IROS}. However, most current line segment detection methods model line segments as straight lines, thus cannot be directly applied to the distorted images from fisheye 
or spherical cameras, which are widely used in indoor camera localization \cite{IndoorLocalization1, IndoorLocalization2}, room layout estimation \cite{Layout1, Layout2}, and other tasks.

The existing methods for distorted line segment detection are almost model-based, i.e., dependent on camera distortion parameters. One category of methods among them requires rectification of lens distortion before applying straight line segments detector. The other methods such as extended Hough transform \cite{FisheyeRectification1} and RANSAC-based methods \cite{FisheyeRectification2}, utilize camera distortion parameters to model the distorted line segments and can be directly applied to the distorted images. However, the performance of model-based methods is largely dependent on the accuracy of camera distortion parameters, which might be even unavailable in hand.

Therefore, the model-free approach for distorted line segment detection, i.e., independent of camera distortion parameters, is applaudable in practice. As a kind of model-free representation, two endpoints model has been commonly used in straight line segments detectors \cite{AFMPAMI,LCNN,HAWP}, but is not enough to represent curved line segments in distorted images. Considering the strong fitting ability of the Bezier curve which has been successfully applied to arbitrarily-shaped text detection \cite{ABCNet}, we adopt the Bezier curve as a unified representation for line segments in both distorted and undistorted images. Similar to the two endpoints model, the Bezier curve can be represented as a vector parameterized by its equipartition points. Analogically, the line segment detection based on the Bezier curve model can be efficiently tackled with the coordinate regression for equipartition points, and the line classification which can be implemented by the center point detection of the target line segment \cite{CenterNet}.

\begin{figure}[t]
	\begin{center}
		\includegraphics[width=0.85\linewidth]{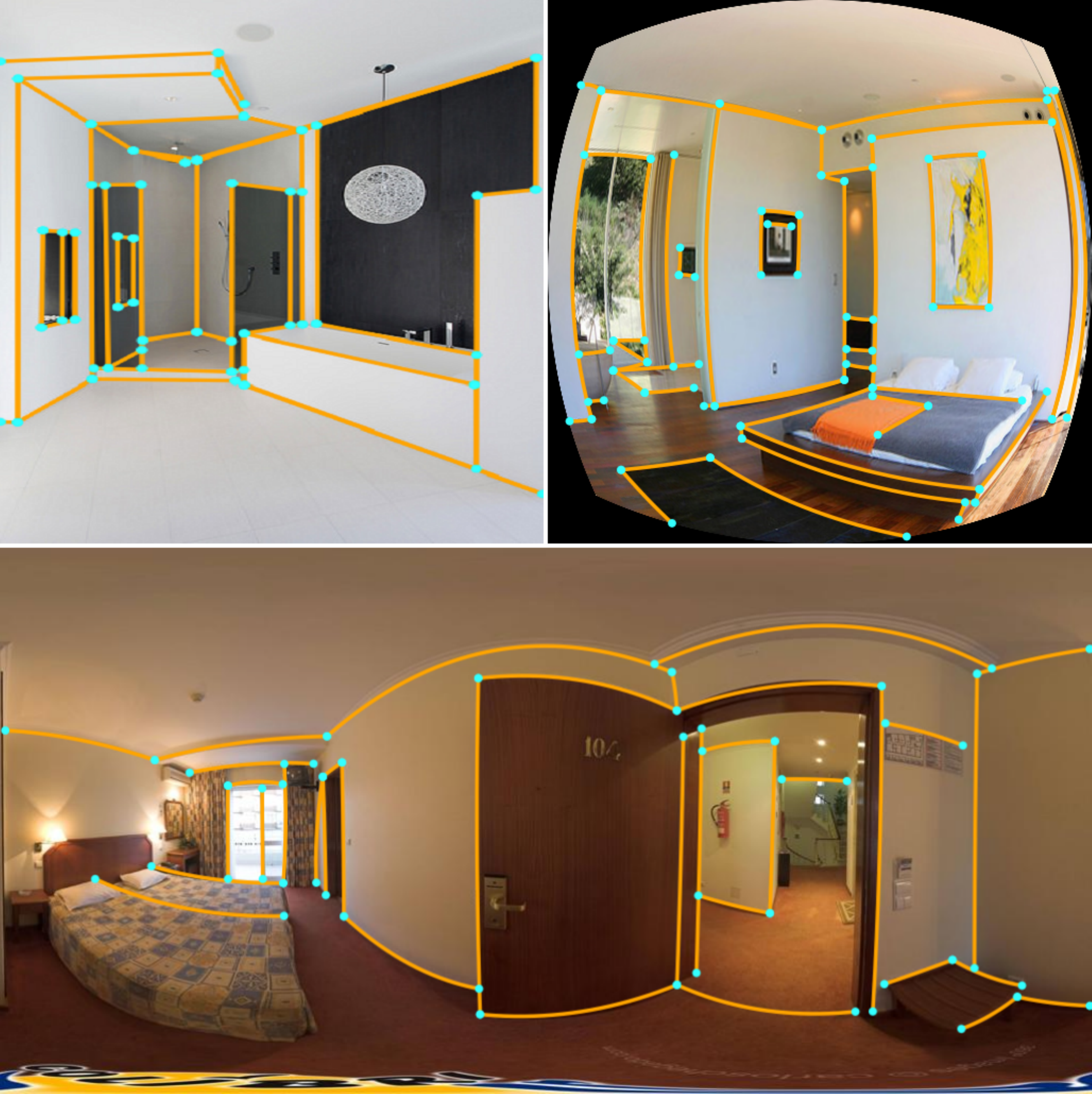}
	\end{center}
	\caption{Demonstration of the line segment detection on a pinhole (top-left), fisheye (top-right), and spherical image (bottom) with the proposed ULSD method.}
	\label{fig:1}
\end{figure}

Exploiting the Bezier curve model and detection method based on the center point, we design an end-to-end network for unified line segment detection (ULSD). The proposed ULSD is a model-free approach that can take both distorted and undistorted images from the pinhole, fisheye or spherical cameras as input, and directly output vectorized line segments, as shown in Fig. \ref{fig:1}. The performance of ULSD is evaluated on various datasets which demonstrates the superiority to state-of-the-art (SOTA). As far as we know, the proposed ULSD is the first deep learning-based method to unify the line segment detection for both distorted and undistorted images. The main contribution of this work is three folded:
\begin{itemize}
	\item We propose a model-free line segment representation for unified line segment detection based on the Bezier curve model, which is independent of camera distortion parameters.
	\item We design a unified end-to-end line segment detection network, which can be directly applied to both distorted and undistorted images from the pinhole, fisheye, and spherical cameras.
	\item We construct fisheye and spherical image datasets for line segment detection tasks in distorted images. With these datasets, the performance of the proposed ULSD is evaluated.
\end{itemize}


\section{Related Work} \label{sec2}
Line segment detection is an attractive research topic during the last two decades in computer vision and robotics. Most of the related methods work well for undistorted pinhole images. For distorted images from fisheye or spherical cameras, the common practice is to undistort images and then deploy the straight-line models. Thus we will mainly review the straight line detection and distorted line detection methods.   

\subsection{Straight Line Segment Detection} \label{sec2-1}
For pinhole images, a lot of line segment detection methods have been proposed. Among them, traditional approaches such as \cite{HT,PHT,LSD} detect lines based on the edge or gradient. The main drawback is that they are sensitive to noise and the detected lines are often fragmented. Recently, deep learning-based works such as \cite{AFM, PPGnet} significantly improve the performance by leveraging the deep features. Compared to the local edge or gradient features, the learning-based features are more robust to noise. Huang et al. \cite{DWP} propose the wireframe representation and provide the first high-quality wireframe dataset. Compared with traditional line segment representation, wireframe parsing leverages the constraint of endpoint junctions, thus the output line segments are of higher quality in terms of line completeness and robustness to noise. Their method detects wireframe by two deep neural networks combined with a heuristic wireframe fusion algorithm. Zhou et al. propose the first end-to-end trainable neural network named L-CNN \cite{LCNN} for wireframe parsing. Based on L-CNN, Xue et al. introduce a holistic attraction field map (HAFM) \cite{HAWP} to represent line segments and achieve SOTA performance in accuracy and efficiency. Although HAFM brings significant improvement of performance, it needs big efforts to expand from the straight line to the distorted line.

\subsection{Distorted Line Segment Detection} \label{sec2-2}
As far as we know, there is no specially-designed method to detect distorted line segments for fisheye or spherical images. But there are some domains closely related. In the straight line-based fisheye image rectification field, various distorted lines detection methods have been used. These approaches can be divided into geometric-based methods \cite{FisheyeRectification1, FisheyeRectification2} and deep learning-based  methods \cite{FisheyeRectification3}. In \cite{FisheyeRectification1}, an extended Hough transform is utilized to detect the straight lines. The work \cite{FisheyeRectification2} proposes a $5$-points RANSAC method for robust line extraction. By leveraging the strong capability of networks, LaRecNet \cite{FisheyeRectification3} can obtain more accurate results of distorted lines extraction. As for spherical images, there are some related works about line-based spherical camera localization \cite{IndoorLocalization1, IndoorLocalization2}. In \cite{IndoorLocalization2}, the authors first utilize Canny edge detection in the 2D equirectangular image, and then deploy spherical Hough transform to extract lines. In general, most of the above works are based on the traditional Hough transform or RANSAC algorithm. These methods are generally sensitive to noise and their detection accuracies are far from satisfactory. Additionally, there is no existing learning-based method modeling the distorted line segments in fisheye or spherical images. 

\section{Overview of the proposed method} \label{sec3}

\begin{figure}[t]
	\begin{center}
		\includegraphics[width=0.99\linewidth]{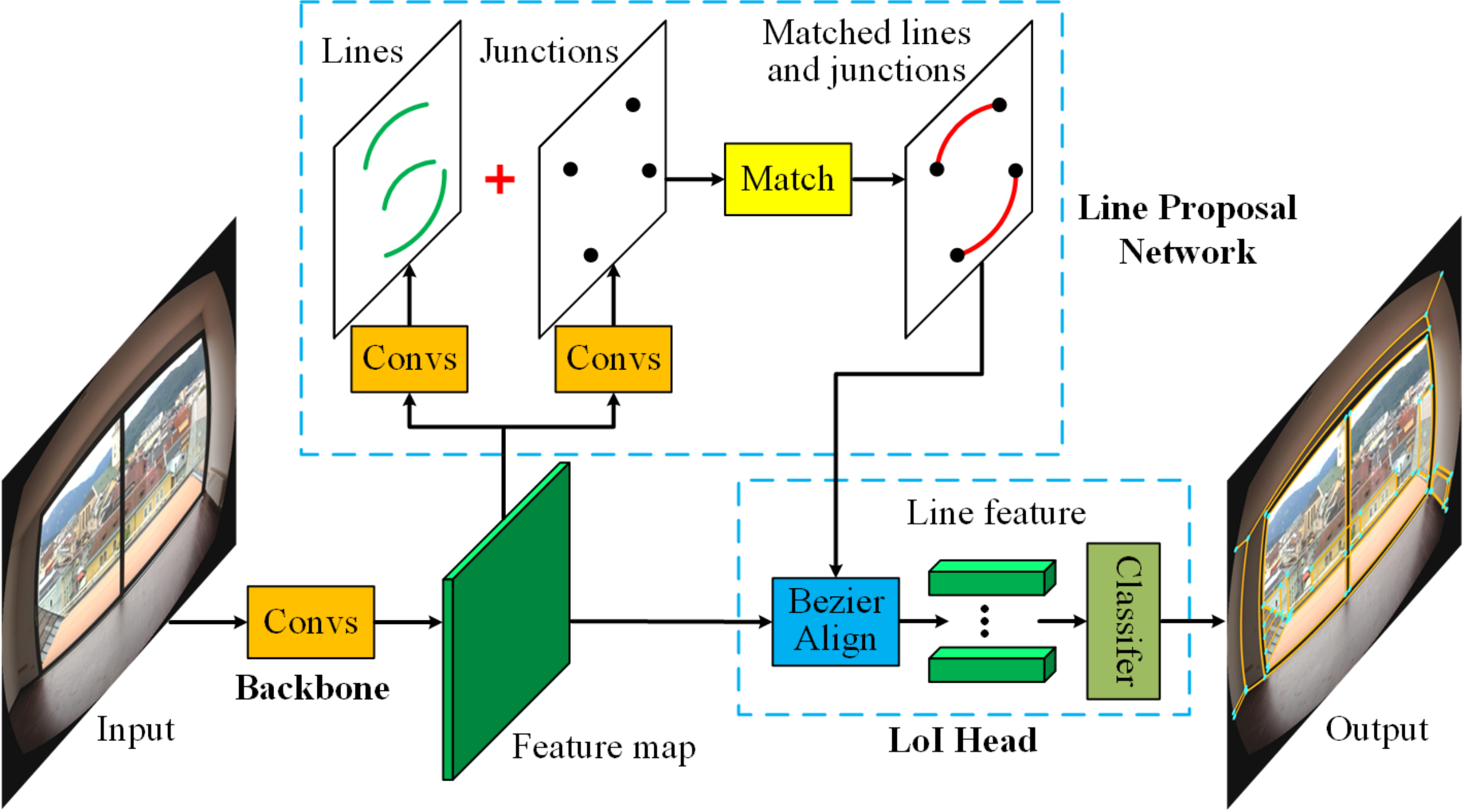}
	\end{center}
	\caption{An overview of our network architecture.}
	\label{fig:2}
\end{figure}

\subsection{Bezier Curve Representation} \label{sec3-1}
Since the most general line representation with the straight connection of two endpoints cannot fit lines in arbitrarily distorted images, we introduce the Bezier curve as a unified parameterized representation. The Bezier curve uses the Bernstein Polynomials as its basis to represent a parametric curve. Its definition is shown in Eq. \ref{eq1}.
\begin{equation} \label{eq1}
	B(t)=\sum \nolimits_{i=0}^{n} b_{i} B_{i, n}(t), 0 \leq t \leq 1
\end{equation}
where $t$ is the proportional coefficient of a point on the curve, $n$ represents the order of the Bezier curve, $b_i$ represents the $i$-th control point, and $B_{i,n}(t)=C_n^i t^i (1-t)^{n-i}$ represents the Bernstein basis Polynomial.

According to Eq. \ref{eq1}, the interpolation formula for a Bezier curve can be obtained, as shown in Eq. \ref{eq2}:
\begin{equation} \label{eq2} 
	\left[\begin{array}{ccc}
		B_{0,n}\left(t_{0}\right) \!&\! \cdots \!&\! B_{n,n}\left(t_{0}\right) \\
		B_{0,n}\left(t_{1}\right) \!&\! \cdots \!&\! B_{n,n}\left(t_{1}\right) \\
		\vdots \!&\! \ddots \!&\! \vdots \\
		B_{0,n}\left(t_{m-1}\right) \!&\! \cdots \!&\! B_{n,n}\left(t_{m-1}\right)
	\end{array}\right]
	\left[\begin{array}{c}
		b_{0} \\
		b_{1} \\
		\vdots  \\
		b_{n}
	\end{array}\right]
	= \\
	\left[\begin{array}{c}
		p_{0} \\
		p_{1} \\
		\vdots  \\
		p_{m-1}
	\end{array}\right]
\end{equation}
where $m$ is the number of the interpolation points, $p_i$ represents the $i$-th interpolation point.

\begin{figure}[h]
	\begin{center}
		\includegraphics[width=0.80\linewidth]{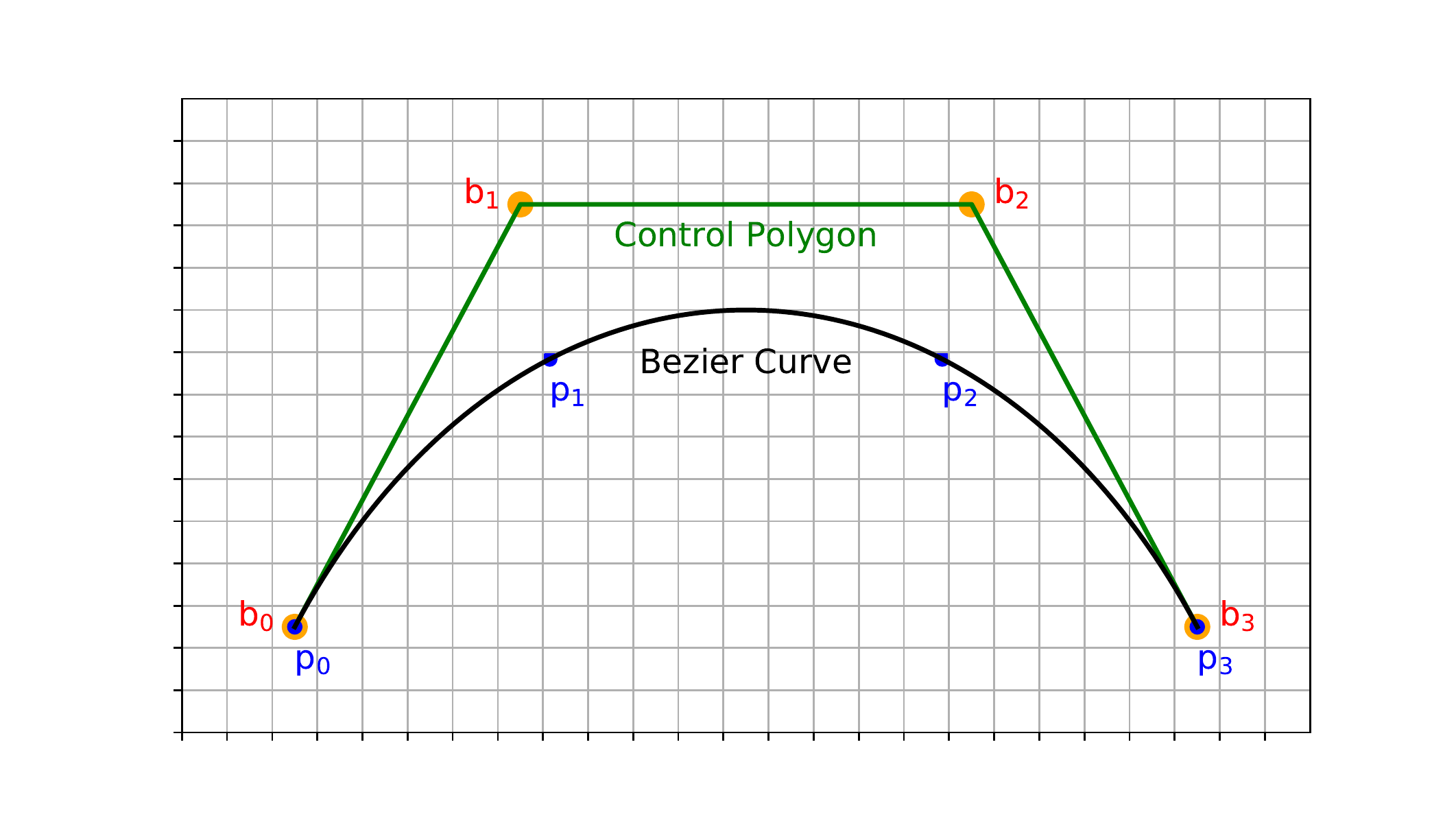}
	\end{center}
	\caption{A $3$rd order Bezier curve. $b_i$ represents the control points, and $p_i$ represents the equipartition points.}
	\label{fig:3}
\end{figure}

An $n$-th order Bezier curve can be determined by its $n+1$ control points. However, the control points lack geometric meaning and may be located outside the image, so it's difficult to directly learn the positions of the control points. Therefore, our network tries to predict the positions of the equipartition points of the Bezier curve instead, and then use Eq. \ref{eq2} to calculate the control points through the least square method. As shown in Fig. \ref{fig:3}, a $3$rd Bezier curve can be determined by $4$ control points $b_i$, but the prediction of $b_1, b_2$ is non-trivial. Therefore, we represent the Bezier curve by using its $n+1$ equipartition points $p_0, p_1, p_2, p_3$.

\subsection{Overall Network Architecture} \label{sec3-2}
Based on the Bezier curve representation, our network is designed to detect line segments in arbitrarily distorted images (from the pinhole, fisheye, and spherical camera). Fig. \ref{fig:2} illustrates the network architecture. It mainly contains three modules: 1) a feature extraction backbone that takes a single image as input and outputs a shared feature map for the successive modules; 2) a Line Proposal Network (LPN) which outputs the candidate line segments; 3) an LoI (Line of Interest) head module which classifies the candidate line segments using the line features obtained through the BezierAlign module. Our pipeline is similar to the Faster R-CNN \cite{FasterRCNN}.

\subsection{Backbone} \label{sec3-3}
We choose the stacked hourglass network \cite{SHN} as the backbone for its efficiency and effectiveness. Taking an image with size $H \times W \times 3$ as input, the stacked hourglass network first downsamples the input image via convolution layers, then extracts features through multiple hourglass modules, and finally outputs the feature map with size $H_b \times W_b \times C$. The feature map is shared by the subsequent LPN and LoI head.

\subsection{Line Proposal Network} \label{sec3-4}
The Line Proposal Network contains four sub-modules: junction prediction module, line prediction module, line and jucntion matching module, and line sample module.

\subsubsection{\textbf{Junction Prediction Module}} \label{sec3-4-1}
Junction prediction is addressed as a classification and regression problem. The input image with spatial size $H \times W$ is divided into $W_b \times H_b$ bins, same as the spatial size of the feature map. For each bin $b$, the network predicts whether there exists a junction inside it. If a junction $\boldsymbol{p}$ is inside bin $b$, it will also predict the offset vector from $\boldsymbol{p}$ to the center $\boldsymbol{b}$ of the bin. Therefore, the network outputs a junction confidence map $J$ and a junction offset map $\boldsymbol{O}$. The ground truth of the two maps can be obtained by Eq. \ref{eq3} and Eq. \ref{eq4}.
\begin{equation} \label{eq3}
	J(b)=\left\{\begin{array}{cl}
		1 & \exists \ \boldsymbol{p} \in {V}: \boldsymbol{p} \ \text{inside} \ b \\
		0 & \text { otherwise }
	\end{array}\right.
\end{equation}
\begin{equation} \label{eq4}
	\mathbf{O}(b)=\left\{\begin{array}{cl}
		\left(\boldsymbol{b}-\boldsymbol{p}\right) & \exists \ \boldsymbol{p} \in {V}: \boldsymbol{p} \ \text{inside} \ b \\
		\boldsymbol{0} & \text { otherwise. }
	\end{array}\right.
\end{equation}
where $V$ is the set of junctions.

$J$ and $\boldsymbol{O}$ are predicted by two decoder head consisting of two convolution layers separately. In the training phase, the binary cross-entropy loss and smooth $l_1$ loss are used to predict $J$ and $\boldsymbol{O}$ respectively. The total loss of junction prediction is the weighted sum of the two losses.
\begin{equation} \label{eq5}
	L_{junc}=\lambda_{conf}^{j} L_{conf}^{j}+\lambda_{offset}^{j} L_{offset}^{j}
\end{equation}

Furthermore, the non-maximum suppression (NMS) is applied in $J$ to remove duplicates, and only the top-$K$ junctions with the highest confidence are kept for the line and junction matching module.

\subsubsection{\textbf{Line Prediction Module}} \label{sec3-4-2}
For an arbitrarily distorted line segment represented by an $n$-th order Bezier curve, the line prediction module tries to predict the location of the center point of the line segment and the offset vectors from the equipartition points to the center point. Center point prediction is the same as junction prediction. If $n$ is even, the center point is one of the $n+1$ equipartition points, whose offset vector is $\boldsymbol{0}$, so only $n$ offset vectors need to predict. The smooth $l_1$ loss is used to predict the offset vectors, and the total loss is shown in Eq. \ref{eq6}.
\begin{equation} \label{eq6}
	L_{line}=\lambda_{center} L_{center}+\lambda_{offset} \sum \nolimits_{i=1}^{m} L_{offset}
\end{equation}
where $m=n$ if $n$ is even, otherwise $m=n+1$.

\subsubsection{\textbf{Line and Junction Matching Module}} \label{sec3-4-3}
To improve the quality of the line segment proposals, a line and junction matching module is adopted. The matching strategy is similar to HAWP \cite{HAWP}. A line segment proposal is kept if and only if its two endpoints can be matched with two junction proposals based on the Euclidean distance, and then the two endpoints of the line proposals are replaced by the two matched junction proposals. If there are multiple line proposals matched with the same pair of junction proposals, only the one with the shortest distance is kept.

\subsubsection{\textbf{Line Sample Module}} \label{sec3-4-4} 
The Line sample module is used to sample positive and negative line proposals for training the classifier of the LoI head. A line segment proposal is assigned with a positive label if there is a ground truth line segment and their distance calculated by Eq. \ref{eq7} is less than a predefined threshold $\eta$. Otherwise, it's assigned with a negative label. Then, we can obtain a positive and a negative sample set. Finally, a certain number of positive and negative line segment proposals are randomly sampled from the two sets respectively. Apart from the above positive samples, we also sample some positive line segments from the ground truth to increase the number of positive samples and help cold-start \cite{LCNN} the training at the beginning.
\begin{equation} \label{eq7}
	\begin{array}{c}
		d(l,l^{\prime}) = \min \left(
		\sum_{i=0}^{n}\left\| \boldsymbol{p}_{i}-\boldsymbol{p}_{i}^{\prime} \right\|^{2},
		\sum_{i=0}^{n}\left\| \boldsymbol{p}_{i}-\boldsymbol{p}_{n-i}^{\prime} \right\|^{2} 
		\right), \\
		l = (\boldsymbol{p}_0,\cdots,\boldsymbol{p}_n), 
		l^{\prime} = (\boldsymbol{p}_0^{\prime},\cdots,\boldsymbol{p}_n^{\prime})
	\end{array}
\end{equation}

\subsection{LoI Head Module} \label{sec3-5}
The LoI (Line of Interest) head module takes a list of candidate line segments together with the feature map $F$ as input and predicts whether or not each candidate line segment is true. To extract the fixed-length line feature vector from the feature map, previous methods \cite{LCNN, HAWP} adopt LoI Pooling layer \cite{LCNN} which is based on the linear interpolation of straight line segments. However, LoI Pooling does not work for the distorted line segments. Therefore, we introduce the BezierAlign layer. Based on Eq. \ref{eq2}, it can uniformly sample $N_p$ points from a distorted line segment. The feature for each sampled point is computed from $F$ using bi-linear interpolation. After a $1D$ max-pooling operator, all the features from the $N_p$ points are concatenated as the line feature vector. After the BezierAlign operation, we feed all feature vectors into a classifier consisting of multiple fully-connected layers followed by a sigmoid layer, and finally get the confidence of each candidate line segment. The binary-cross entropy loss is used in the LoI head module. To balance the loss of positive and negative samples, the two losses are calculated and weighted separately. The total loss of the LoI head module is shown in Eq. \ref{eq8}.
\begin{equation} \label{eq8}
	L_{cls}=\lambda_{pos} L_{pos}+\lambda_{neg} L_{neg}
\end{equation}

The loss of the whole network is the sum of all above losses.
\begin{equation} \label{eq9} 
	L=L_{junc}+L_{line}+L_{cls}
\end{equation}

\section{Experiments} \label{sec4} 
In this section, we will discuss the datasets, network implementation details, and the experimental results. 

\subsection{Datasets} \label{sec4-1} 
\noindent \textbf{Pinhole image}. The common Wireframe dataset \cite{DWP} and YorkUrban dataset \cite{YorkUrban} are used. The former contains 5000 training images and 462 testing images. The latter contains 102 images. 

\noindent \textbf{Fisheye image}. Currently, there is no available public fisheye line segment detection dataset, thus we build the F-Wireframe dataset and F-YorkUrban dataset by distorting images from the Wireframe dataset and YorkUrban dataset with the fisheye distortion model. The synthesized dataset is the same size as the original dataset. 

\noindent \textbf{Spherical image}. Since no public spherical line segment dataset is available, we build a dataset by manually annotating images from the SUN360 dataset \cite{SUN360} and then augmenting by flip (horizontal, vertical, and horizontal-vertical) and periodical shifting (horizontal). Finally, the dataset contains 5200 training images and 68 testing images.

\subsection{Implementation Details} \label{sec4-2} 
For the pinhole and fisheye networks, the input image is resized to $(H,W)=(512,512)$ and the feature map spatial size is $(H_b,W_b)=(128,128)$. For the spherical network, the input image size and the feature map spatial size are $(512,1024)$ and $(128,256)$ respectively. The network's hyper-parameter settings are the same as L-CNN and HAWP for a fair comparison. Our network is trained using the Adam optimizer \cite{Adam}. The learning rate, weight decay, and training batch size are set to $4 \times 10^{-4}$, $1 \times 10^{-4}$, and $6$ respectively. We use step decay as the learning rate scheduler. All experiments are conducted on a single NVIDIA GTX 2080Ti GPU.

\begin{figure}[t]
	\begin{center}
		\includegraphics[width=0.80\linewidth]{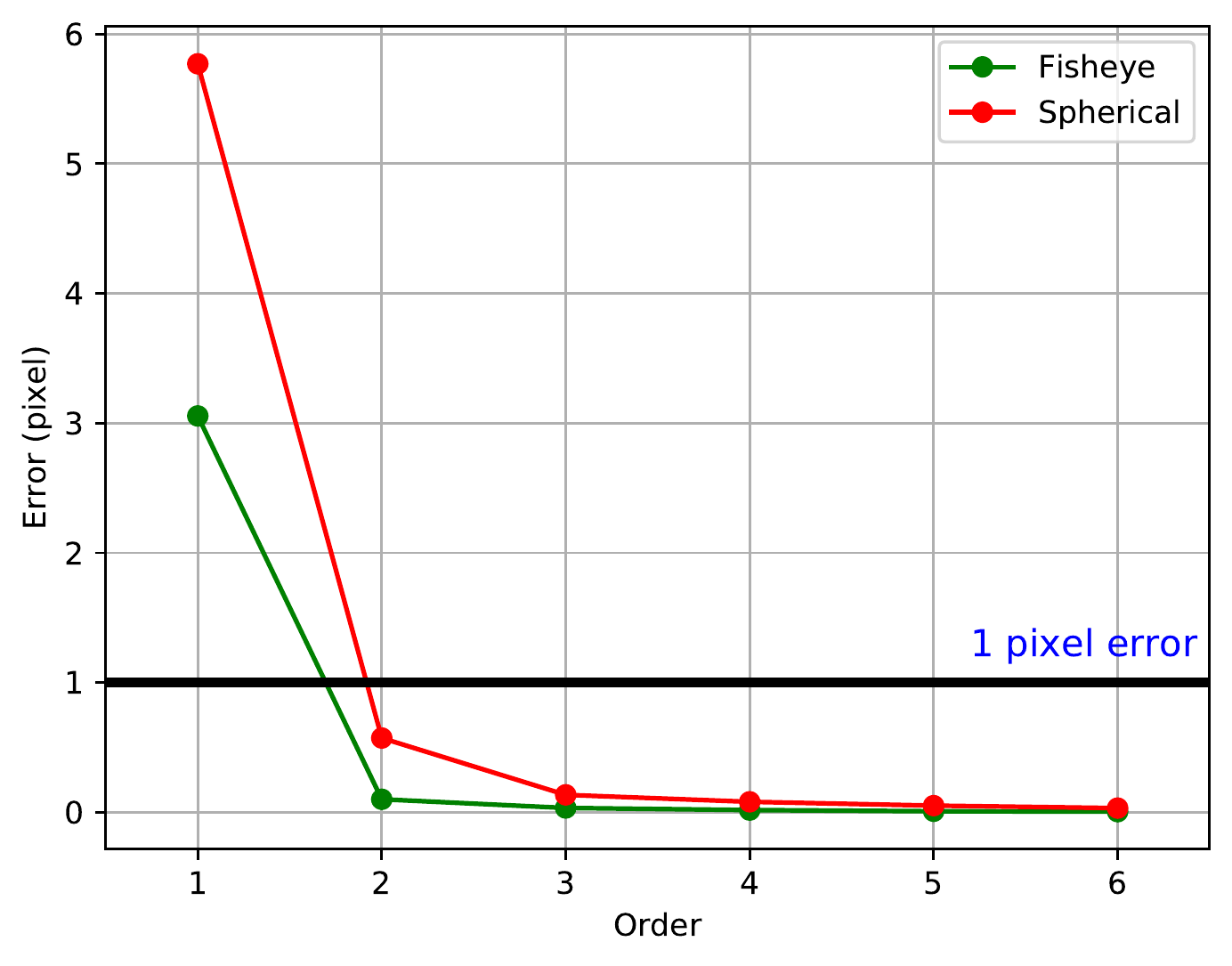}
	\end{center}
	\caption{Illustration of the fitting errors of Bezier curves with different orders for fisheye and spherical images.}
	\label{fig:4}
\end{figure}

As for the order setting of the Bezier curve, the $1$st order is enough for pinhole images. While for fisheye and spherical images, the order is determined by experimental evaluations. We measure the fitting errors of Bezier curves with the order ranging from the $1$ to $6$. As shown in Fig. \ref{fig:4}, the $2$nd order Bezier curve is enough to make the fitting error less than $1$ pixel for line segments in both spherical and fisheye images. Even though, the fitting error can be further decreased by increasing the order, this will largely increase the computational complexity. To balance the accuracy of line representation and efficiency of the network, we leave the setting for the order of the Bezier curve model in the following section by evaluating the performance of ULSD with the order of $2$ to $4$ for fisheye and spherical images.

\subsection{Results and Comparisons} \label{sec4-3} 
To evaluate the performance of the proposed ULSD, comparisons are made to the conventional methods LSD \cite{LSD} and spherical Hough transform (SHT), deep learning-based methods DWP \cite{DWP}, AFM \cite{AFM}, L-CNN \cite{LCNN}, and HAWP \cite{HAWP}. For pinhole images targeting to detect straight line segments, the USLD with 1st order Bezier curve (ULSD$^1$) is trained on the Wireframe dataset and tested on both two pinhole image datasets. For fisheye and spherical images with distorted line segments, the ULSD with the 2nd, 3rd, and 4th order Bezier curve (ULSD$^2$, ULSD$^3$, ULSD$^4$) are respectively evaluated by training on the F-Wireframe dataset and SUN360 dataset.

\begin{table}[h]
	\caption{{\color{black}Quantitative results and comparisons on the Wireframe dataset and YorkUrban dataset.}}
	\centering
	\setlength{\tabcolsep}{1.0mm}{
		\small
		\begin{tabular}{c|ccc|ccc|c}
			\hline
			\multirow{2}{*}{Method} & \multicolumn{3}{c|}{\text{Wireframe Dataset}} & \multicolumn{3}{c|}{\text{YorkUrban Dataset}} & \multirow{2}{*}{FPS} \\ 
			& sAP$^{10}$ & msAP & mAP$^{J}$ & sAP$^{10}$ & msAP & mAP$^{J}$ & \\
			\hline
			LSD \cite{LSD} & 9.5 & 9.3 & 17.2 & 9.4 & 9.4 & 15.4 & \textbf{50.9} \\
			\hline
			DWP \cite{DWP} & 6.8 & 6.6 & 38.6 & 2.7 & 2.7 & 23.4 & 2.3 \\
			AFM \cite{AFM} & 24.3 & 23.4 & 24.3 & 9.1 & 8.9 & 12.5 & 14.3 \\	
			L-CNN \cite{LCNN} & 62.9 & 62.1 & 59.3 & 26.4 & 26.1 & 30.4 & 13.7 \\	
			HAWP \cite{HAWP} & \textbf{66.5} & \textbf{65.7} & 60.2 & \textbf{28.5} & \textbf{28.1} & \textbf{31.7} & 30.9 \\					
			\hline
			ULSD$^1$ (ours) & 66.4 & 65.6 & \textbf{61.4} & 27.4 & 27.0 & 31.0 & \textbf{40.6} \\
			\hline
	\end{tabular}}
	\label{tab:1}
\end{table} 
\begin{figure}[h!] 
	\centering
	\begin{minipage}[t]{0.49\linewidth}
		\centering
		\includegraphics[width = 1\columnwidth]{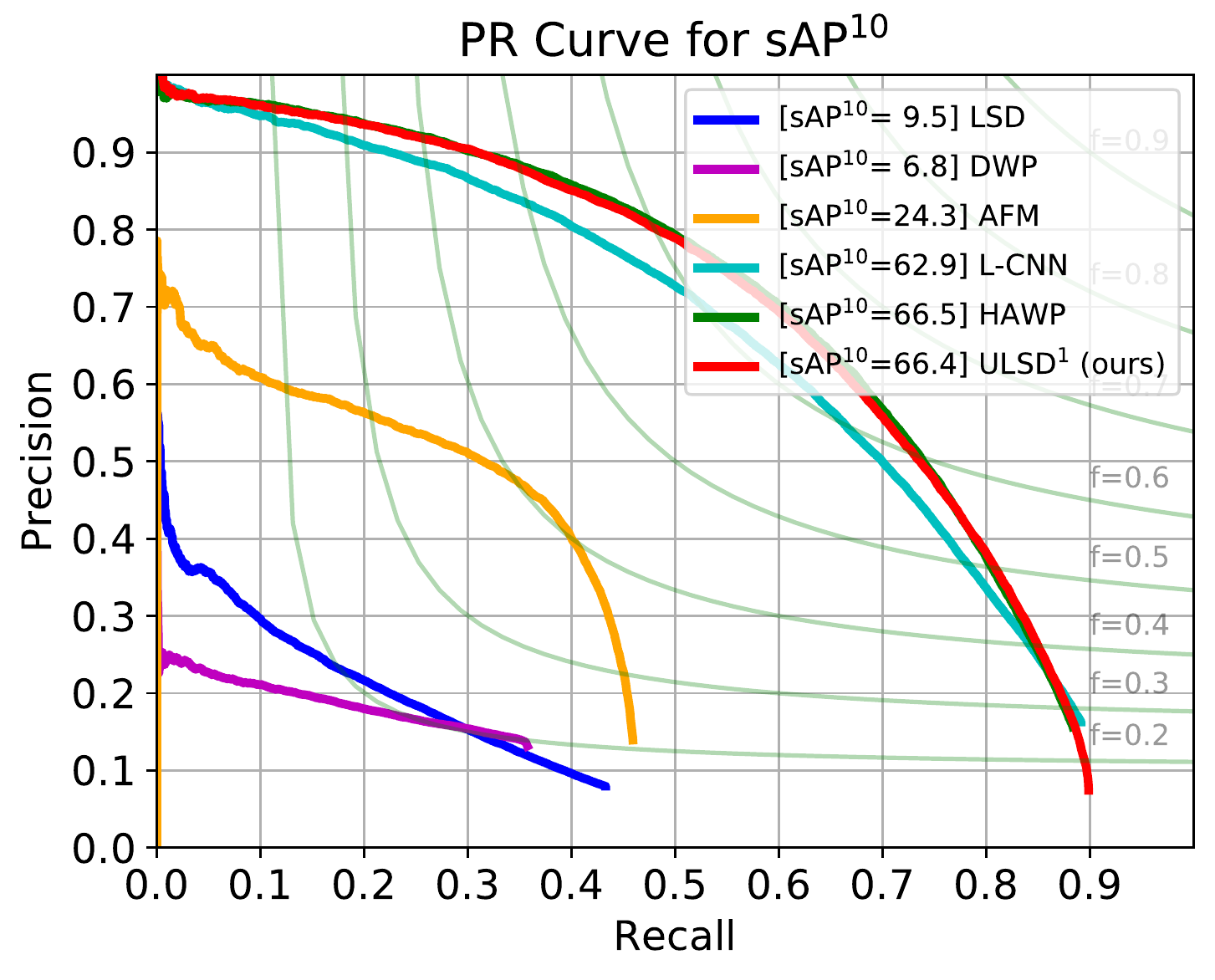} 
		\label{fig:5a}
	\end{minipage}
	\begin{minipage}[t]{0.49\linewidth}
		\centering
		\includegraphics[width = 1\columnwidth]{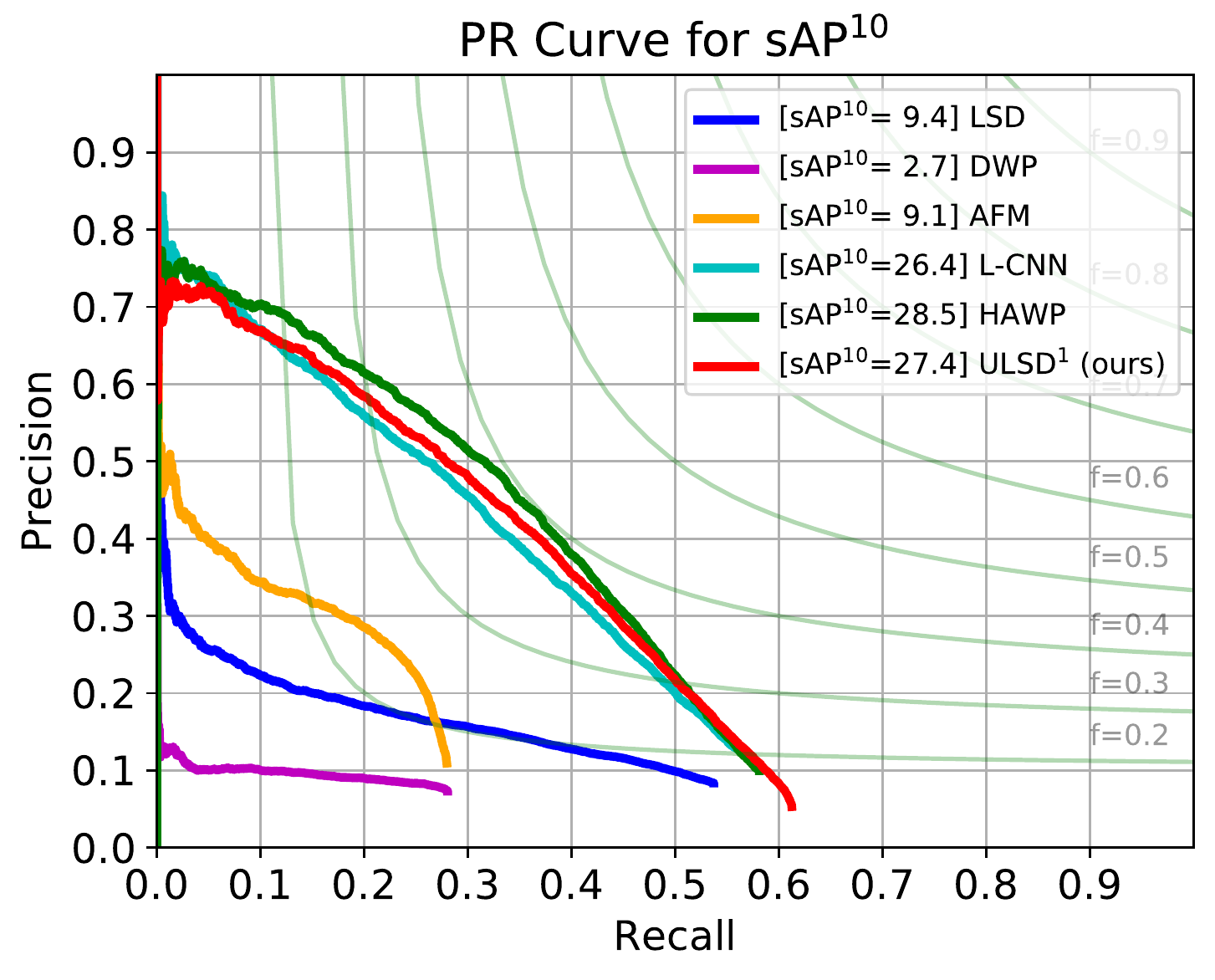}
		\label{fig:5b}
	\end{minipage}
	\caption{Precision-Recall (PR) curves of sAP$^{10}$ on the Wireframe dataset (the left plot) and YorkUrban dataset (the right plot).}
	\label{fig:5}
\end{figure}
\noindent \textbf{Quantitative Results}. Quantitative performance in the accuracy of line segment detection is evaluated based on metrics including structural average precision (sAP) of line segments under the threshold of $5$, $10$, $15$ pixels, mean structural average precision (msAP) over the threshold of $5$, $10$, and $15$ pixels and vectorized junction mean AP (mAP$^J$). And the efficiency of the aforementioned algorithms is further evaluated based on the frames per second (FPS). 

The quantitative results and comparisons for pinhole images are shown in Table. \ref{tab:1} and Fig. \ref{fig:5}. The proposed ULSD$^1$ obtains much higher detection accuracy compared with traditional LSD. Furthermore, compared to the deep learning-based methods, the proposed ULSD achieves the accuracy comparable to the SOTA, but with remarkable improvement in efficiency by at least $31\%$, which comes from the high efficiency of the ULSD's line segment prediction module.

\begin{table}[htbp]
	\caption{{\color{black}Quantitative results and comparisons on the F-Wireframe dataset and F-YorkUrban dataset.}}
	\centering
	\setlength{\tabcolsep}{1.0mm}{
		\small
		\begin{tabular}{c|ccc|ccc|c}
			\hline
			\multirow{2}{*}{Method} & \multicolumn{3}{c|}{\text{F-Wireframe Dataset}} & \multicolumn{3}{c|}{\text{F-YorkUrban Dataset}} & \multirow{2}{*}{FPS} \\ 
			& sAP$^{10}$ & msAP & mAP$^{J}$ & sAP$^{10}$ & msAP & mAP$^{J}$ & \\
			\hline
			LSD \cite{LSD} & 4.3 & 4.3 & 11.1 & 5.2 & 5.1 & 10.7 & \textbf{47.9} \\
			\hline
			L-CNN \cite{LCNN} & 43.4 & 42.9 & 44.2 & 19.9 & 19.6 & 26.4 & 14.3 \\
			HAWP \cite{HAWP} & 46.3 & 45.6 & 43.8 & 21.5 & 21.2 & 26.4 & 31.5 \\
			HAWP$^+$ & 56.4 & 55.4 & - & 25.8 & 25.4 & - & 31.5 \\
			\hline
			ULSD$^2$ (ours) & \textbf{61.2} & \textbf{60.2} & \textbf{56.3} & \textbf{30.2}& \textbf{29.6} & 32.6 & \textbf{36.8} \\
			ULSD$^3$ (ours) & 60.3 & 59.3 & 56.1 & 28.6 & 28.0 & 31.5 & 36.5 \\
			ULSD$^4$ (ours) & 59.9 & 59.0 & \textbf{56.3} & 30.1 & \textbf{29.6} & \textbf{33.1} & 36.3 \\
			\hline
	\end{tabular}}
	\label{tab:2}
\end{table} 

\begin{table}[htbp] 
	\caption{{\color{black}Quantitative results and comparisons on the SUN360 dataset. The input image for L-CNN and HAWP is resized to $512 \times 512$, but ULSD remains the original size $512 \times 1024$, thus the speed of ULSD is a little slower than HAWP's.}}
	\centering
	\setlength{\tabcolsep}{1.0mm}{
		\small
		\begin{tabular}{c|ccccc|c}
			\hline
			\multirow{2}{*}{Method} & \multicolumn{5}{c|}{\text{SUN360 Dataset}} &  \multirow{2}{*}{FPS} \\
			& sAP$^{5}$ & sAP$^{10}$ & sAP$^{15}$ & msAP & mAP$^{J}$ & \\
			\hline
			SHT & 0.9 & 1.7 & 2.5 & 1.7 & 3.4 & 0.05 \\
			\hline
			L-CNN \cite{LCNN} & 39.8 & 42.5 & 43.6 & 42.0 & 34.8 & 12.6 \\
			HAWP \cite{HAWP} & 41.7 & 44.7 & 45.8 & 44.1 & 33.1 & \textbf{25.4} \\
			\hline
			ULSD$^2$ (ours) & \textbf{61.9} & \textbf{67.6} & \textbf{69.8} & \textbf{66.4} & \textbf{47.3} & 24.8 \\
			ULSD$^3$ (ours) & 60.9 & 66.7 & 68.7 & 65.4 & 47.0 & 24.6 \\
			ULSD$^4$ (ours) & 60.3 & 66.1 & 68.0 & 64.8 & \textbf{47.3} & 24.4 \\
			\hline
	\end{tabular}}
	\label{tab:3}
\end{table} 

\begin{figure}[h!] 
	\centering
	\begin{minipage}[t]{0.49\linewidth}
		\centering
		\includegraphics[width = 1\columnwidth]{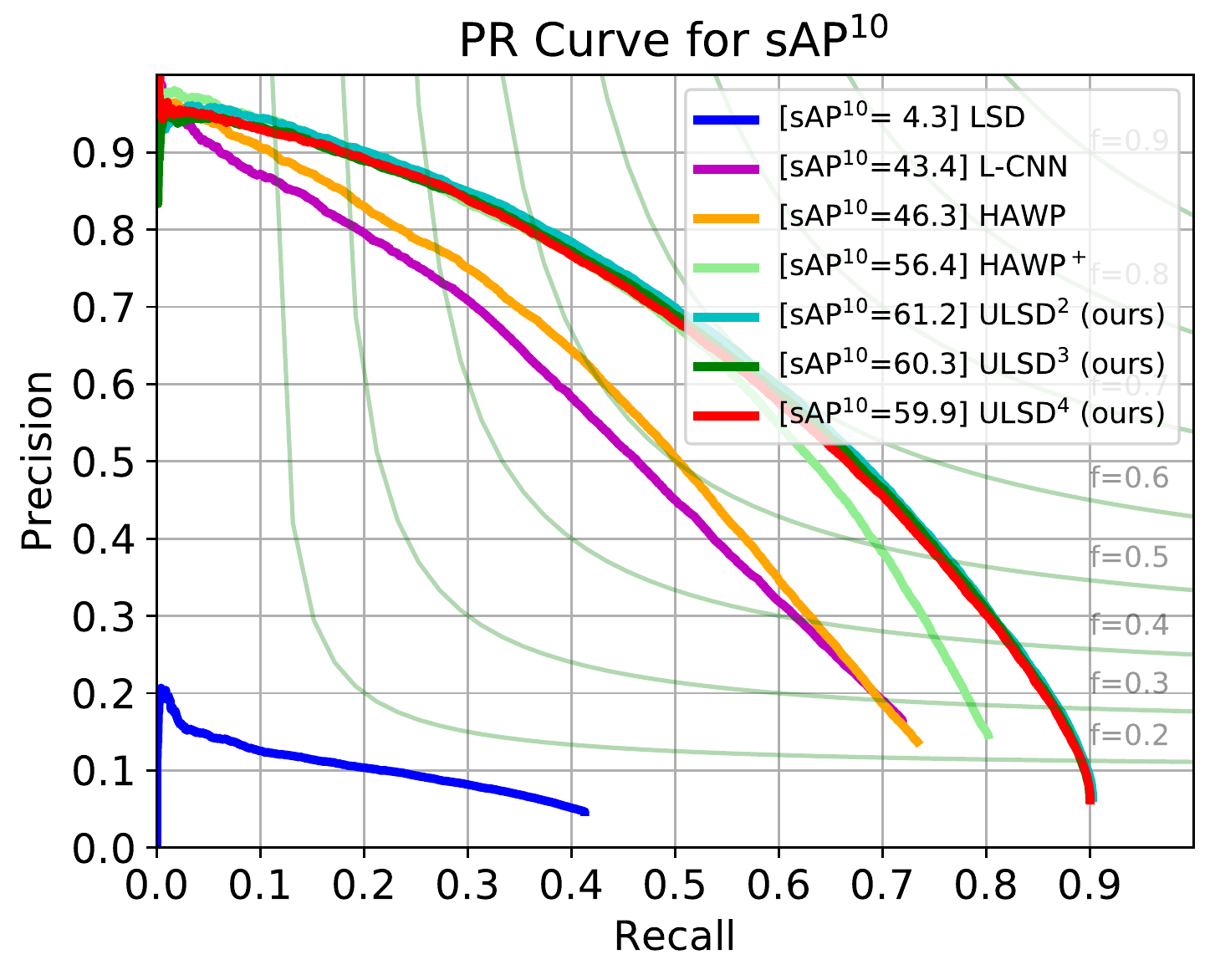} 
		\label{fig:6a}
	\end{minipage}
	\begin{minipage}[t]{0.49\linewidth}
		\centering
		\includegraphics[width = 1\columnwidth]{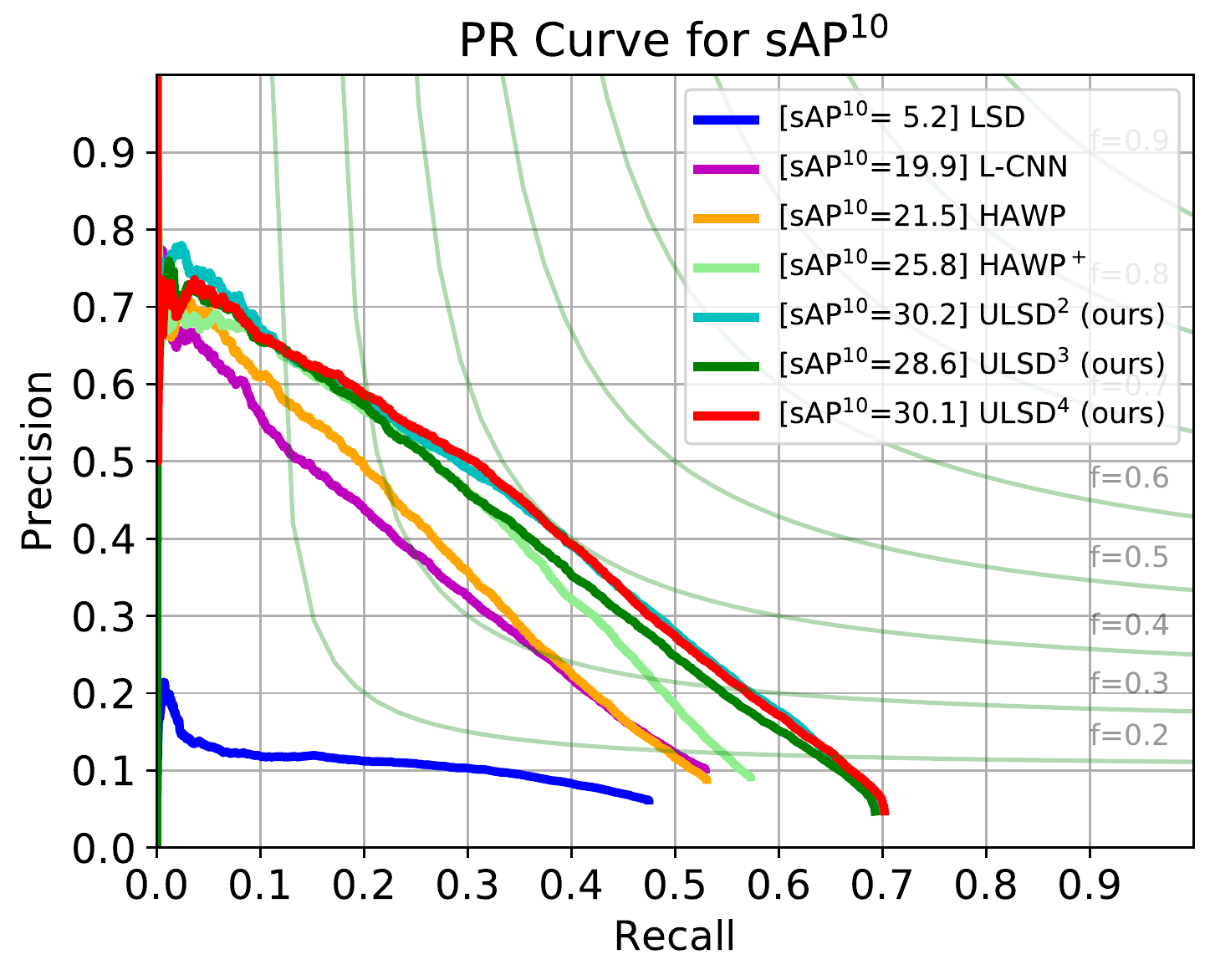}
		\label{fig:6b}
	\end{minipage}
	\caption{PR curves of sAP$^{10}$ on the F-Wireframe dataset (the left plot) and F-YorkUrban dataset (the right plot).}
	\label{fig:6}
\end{figure}

\begin{figure}[h!] 
	\centering
	\begin{center}
		\includegraphics[width=0.5\linewidth]{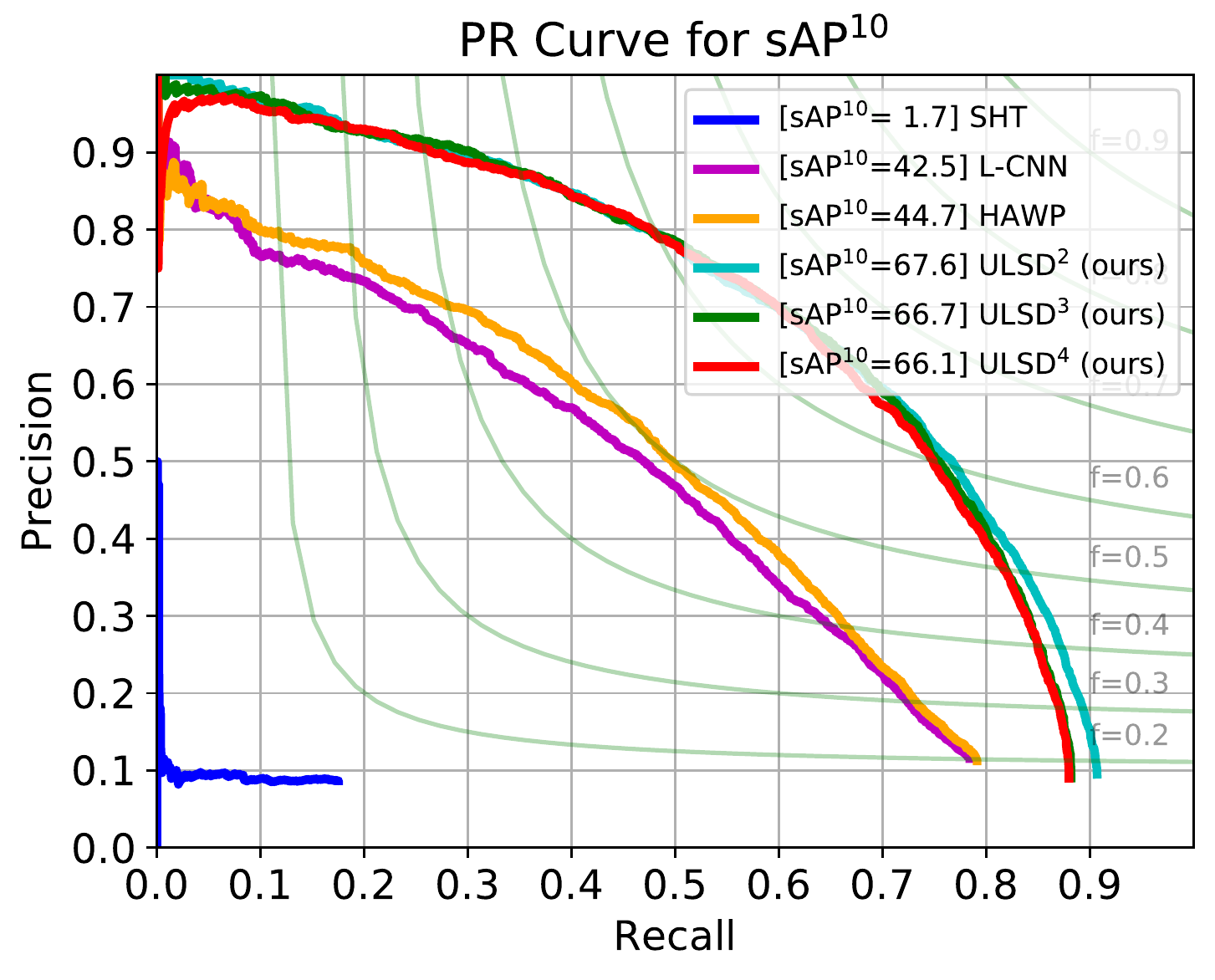}
	\end{center}
	\caption{PR curves of sAP$^{10}$ on the SUN360 dataset.}
	\label{fig:7}
\end{figure}

\begin{figure*}[h!] 
	\centering
	\begin{center}
		\includegraphics[width=0.9\linewidth]{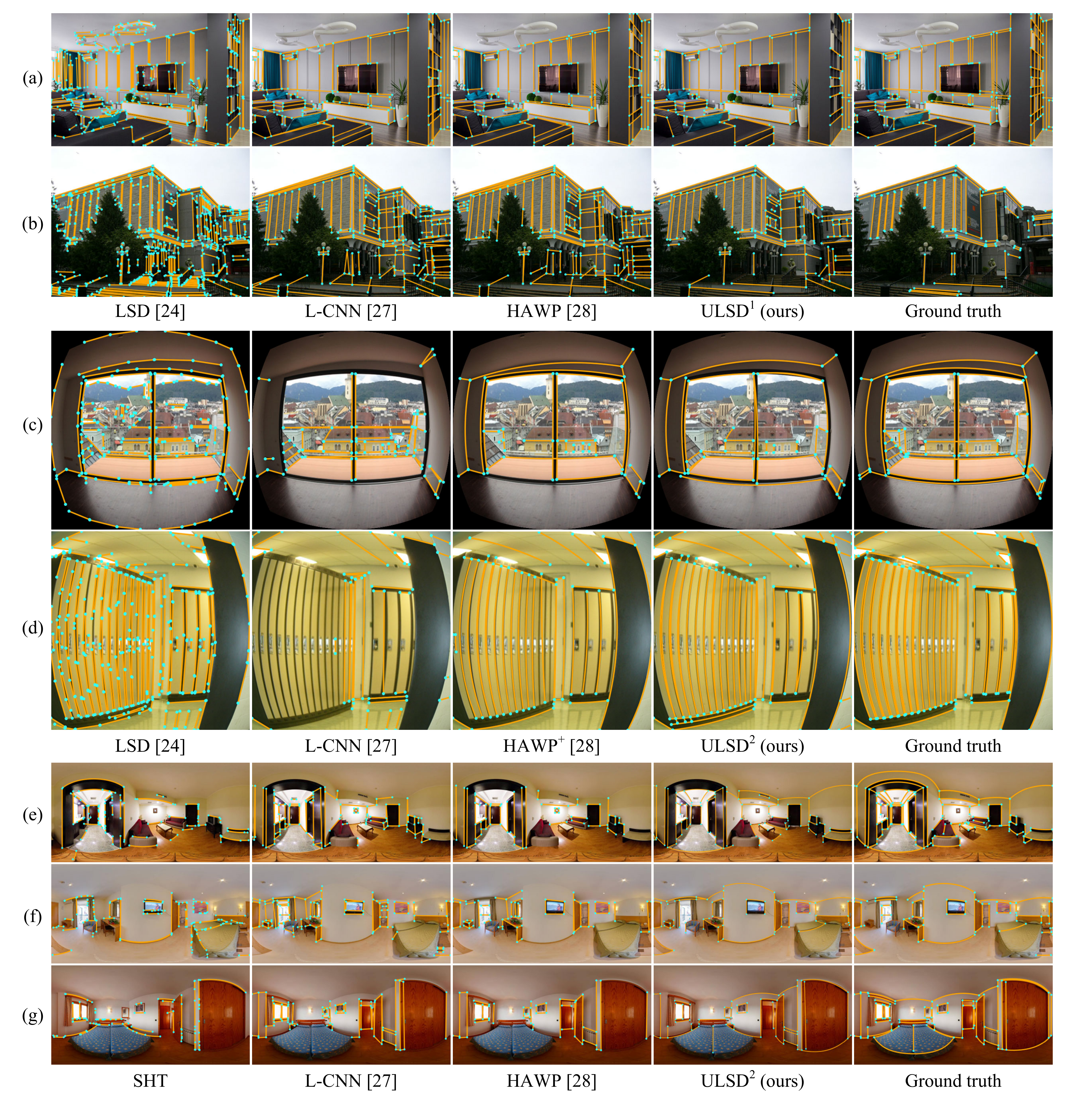}
	\end{center}
	\caption{Qualitative results and comparisons. Rows: (a)-(b) pinhole case, (c)-(d) fisheye case, (e)-(g) spherical case}
	\label{fig:8}
\end{figure*}

Then by deploying higher-order Bezier curve representation, the quantitative performance for different algorithms is evaluated on fisheye and spherical datasets. For the fair comparison, the performance of SOTA straight line segment detection, HAWP, is further evaluated with an additional prepossessing by rectifying the fisheye images with randomly corrupted camera distortion parameters ($10\%$ random noise). And we denote it as HAWP$^+$. The quantitative results are shown in Table. \ref{tab:2}, Table. \ref{tab:3}, and the corresponding PR curves in Fig. \ref{fig:6} and Fig. \ref{fig:7}. For both fisheye and spherical images, the proposed ULSD outperforms the SOTA in both accuracy and efficiency. It is worth noting that although the rectification brings some performance improvement for HAWP$^+$ comparing to its origin, the proposed ULSD still outperforms HAWP$^+$. Even more, ULSD is a model-free method that is independent of camera distortion parameters.

Comparing the results of ULSD on different order Bezier representations, we can observe that ULSD$^2$ is slightly better than higher-order ULSD$^3$ and ULSD$^4$ for both accuracy and efficiency. Although higher-order Bezier representation can improve the line fitting precision (Fig. \ref{fig:4}), less than $1$ pixel fitting error does not make big differences for the precision evaluation. Thus the higher-order Bezier curve representation does not effectively improve the line detection accuracy, instead makes the model more complex and difficult to learn, which results in lower accuracy. Thus, we evaluate the qualitative results with ULSD$^2$ for distorted images.

\noindent \textbf{Qualitative Results}. The qualitative results are shown in Fig. \ref{fig:8}. Since LSD and SHT are based on the edge or gradient, they detect some noise edges without geometric meaning. Besides, they also produce a lot of fragmented line segments due to not exploiting the constraint of junctions. By leveraging the learning-based features as well as the constraints of junctions, deep models L-CNN and HAWP can produce high-quality straight line segments detection results for pinhole images. However, restricted by the two-endpoint representation, they are incapable of detecting the distorted line segments for fisheye and spherical images. Even though, the performance of SOTA methods for straight line segments detection can be further improved by rectification preprocessing, our proposed ULSD armed with the parameterized Bezier curve model for line segments has the best performance both in accuracy and efficiency. Furthermore, ULSD can directly extract line segments in distorted or undistorted images

\section{Conclusion} \label{sec5}
In this paper, we present a unified line segment detection method (ULSD) to detect arbitrarily distorted lines in pinhole, fisheye and, spherical images. With the novel Bezier curve representation, our network can formulate arbitrarily distorted line models in the three core modules: backbone, Line Proposal Network, and LoI head. Compared with the SOTA methods L-CNN \cite{LCNN} and HAWP \cite{HAWP}, our ULSD achieves competitive results for pinhole images. More importantly, it obtains much higher line detection accuracy and efficiency for the fisheye and spherical images. With the real-time line detection speed using a single GPU, the proposed ULSD has great potential for online visual tasks such as SLAM and 3D reconstruction. 

\bibliographystyle{IEEEtran}
\bibliography{egbib}
\end{document}